\documentclass{article}

\usepackage{microtype}
\usepackage{graphicx}
\usepackage{subfigure}
\usepackage{booktabs} 

\usepackage{hyperref}

\usepackage[accepted]{icml2024}

\usepackage{bbm}
\usepackage{amsmath}
\usepackage{amssymb}
\usepackage{mathtools}
\usepackage{amsthm}
\usepackage{natbib}

\usepackage[capitalize,noabbrev]{cleveref}

\theoremstyle{plain}

\theoremstyle{definition}

\theoremstyle{remark}

\usepackage[textsize=tiny]{todonotes}

\icmltitlerunning{Diving into Underwater: Segment Anything Model Guided Underwater Salient Instance Segmentation and A Large-scale Dataset}

\usepackage{multirow}

\def\datasetName{USIS10K}
\def\methodName{USIS-SAM}
\def\eg{e.g.}
\def\ie{i.e.}
\def\mAP{mAP}
\def\AP#1{AP$_{#1}$}

\begin{document}

\twocolumn[
\icmltitle{Diving into Underwater: Segment Anything Model Guided Underwater Salient Instance Segmentation and A Large-scale Dataset}



\icmlsetsymbol{equal}{*}
\icmlsetsymbol{cor}{\dag}

\begin{icmlauthorlist}
\icmlauthor{Shijie Lian}{equal,hnu}
\icmlauthor{Ziyi Zhang}{equal,hkust}
\icmlauthor{Hua Li}{cor,hnu,su}
\icmlauthor{Wenjie Li}{hnu}
\icmlauthor{Laurence Tianruo Yang}{hnu,hzu,StFX}
\icmlauthor{Sam Kwong}{lnu}
\icmlauthor{Runmin Cong}{sdu,edu}
\end{icmlauthorlist}

\icmlaffiliation{hnu}{Hainan University}
\icmlaffiliation{hkust}{The Hong Kong University of Science and Technology (Guangzhou)}
\icmlaffiliation{lnu}{LIngnan University, School of data science, Hong Kong}
\icmlaffiliation{hzu}{Huazhong University of Science and Technology}
\icmlaffiliation{sdu}{Shandong University, School of Control Science and Engineering}
\icmlaffiliation{StFX}{St. Francis Xavier University}
\icmlaffiliation{edu}{Key Laboratory of Machine Intelligence and System Control, Ministry of Education, Jinan, China}
\icmlaffiliation{su}{Key Laboratory of New Generation Artificial Intelligence Technology \& Its Interdisciplinary Applications (Southeast University), Ministry of Education, China}

\icmlcorrespondingauthor{Hua Li}{lihua@hainanu.edu.cn}

\icmlkeywords{Machine Learning, ICML}

\vskip 0.3in
]



\printAffiliationsAndNotice{\icmlEqualContribution} 

\begin{abstract}
With the breakthrough of large models, Segment Anything Model (SAM) and its extensions have been attempted to apply in diverse tasks of computer vision.
Underwater salient instance segmentation is a foundational and vital step for various underwater vision tasks, which often suffer from low segmentation accuracy due to the complex underwater circumstances and the adaptive ability of models.
Moreover, the lack of large-scale datasets with pixel-level salient instance annotations has impeded the development of machine learning techniques in this field.
To address these issues, we construct the first large-scale underwater salient instance segmentation dataset (\datasetName), which contains 10,632 underwater images with pixel-level annotations in 7 categories from various underwater scenes.
Then, we propose an Underwater Salient Instance Segmentation architecture based on Segment
Anything Model (\methodName) specifically for the underwater domain. We devise an Underwater Adaptive Visual Transformer (UA-ViT) encoder to incorporate underwater domain visual prompts into the segmentation network. 
We further design an out-of-the-box underwater Salient Feature Prompter Generator (SFPG) to automatically generate salient prompters instead of explicitly providing foreground points or boxes as prompts in SAM.
Comprehensive experimental results show that our \methodName~method can achieve superior performance on \datasetName~datasets compared to the state-of-the-art methods.
Datasets and codes are released on \href{https://github.com/LiamLian0727/USIS10K}{Github}.
\end{abstract}


\section{Introduction}
\label{sec:introduction}
In recent years, fundamental models such as Generative Pre-Trained Transformer (GPT)-4 \cite{achiam2023gpt}, Language Learning with Adaptive Multi-task Architecture (LLaMA) \cite{touvron2023llama}, and Segment Anything Model (SAM) \cite{SAM_2023_ICCV}~have made significant progress and greatly contributed to the advancement of the society. Among them, SAM has recently performed well in many segmentation tasks due to its excellent encoder-decoder transformation framework and large training dataset SA-1B. 
After being fine-tuned or modified appropriately, it has strong potential for segmentation applications in medical images \cite{wu2023medical, huang2024segment}, remote sensing images \cite{chen2023rsprompter}, and other fields.

\begin{figure}[!t]
  \centering
  \includegraphics[width=1\linewidth]{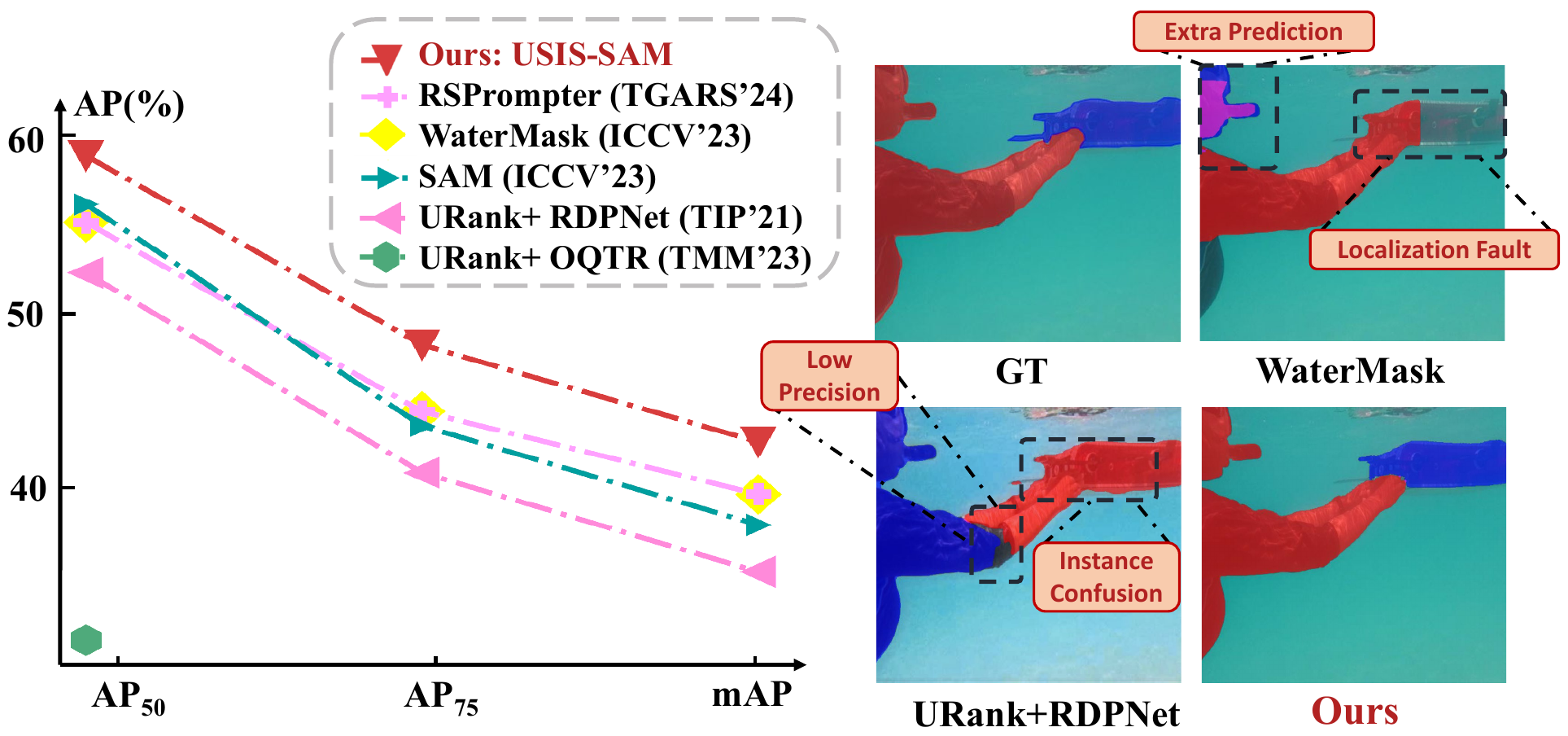}
  \vspace{-5mm}
  \caption{A simple comparison of \methodName~and other state-of-the-art methods trained on the \datasetName~dataset. Different colors represent different salient instances. URank represents the underwater image enhancement method in UnderwaterRanker \cite{underwaterranker_aaai_2023}.}
  \vspace{-5mm}
  \label{fig:simplt_show}
\end{figure}

\begin{figure*}[htbp]
  \centering
  \includegraphics[width=1\linewidth]{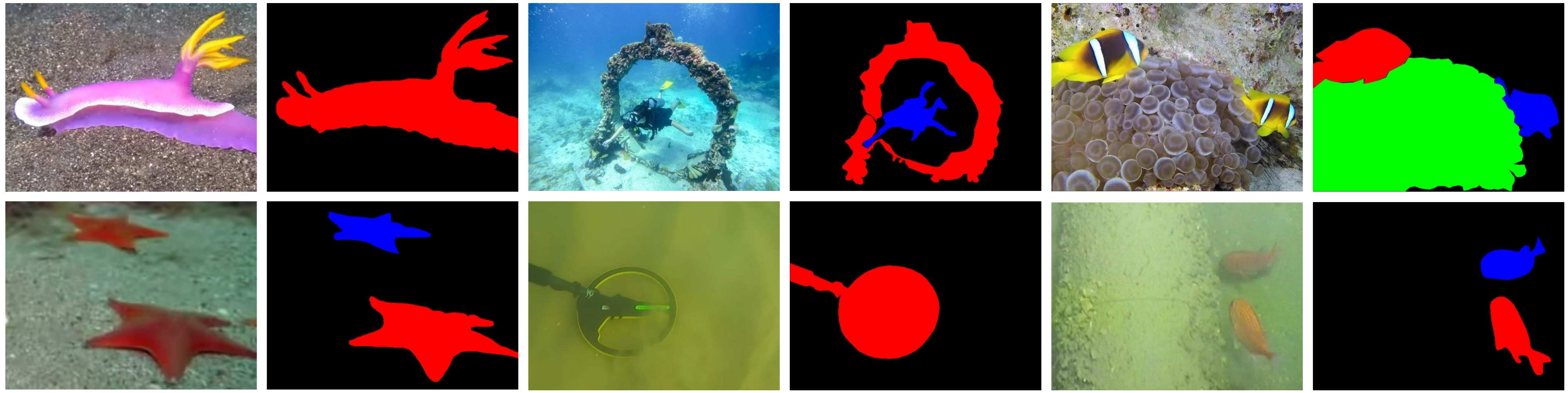}
  \vspace{-6mm}
  \caption{Examples of annotations for various salient instances in \datasetName. The image on the left is the original image and the right is the annotation mask, different colors represent different salient instances. More dataset showings can be found in \Cref{subsec:more_sample_ill}.}
  \vspace{-3mm}
  \label{fig:usis10k_show}
\end{figure*}

To meet the growing demands for underwater exploration, the deep learning vision community is increasingly focusing on the development and application of underwater vision tasks.
Salient Instance Segmentation (SIS), an emerging and promising visual task, aims to segment out visually salient objects in a scene and distinguish individual salient instances \cite{Tian2022-kn}, which is beneficial for vision tasks requiring fine-grained scene understanding. 
Underwater salient instance segmentation (USIS), may benefit for subsequent underwater vision tasks like marine ruins discovery, marine resources exploration, underwater human-robot interaction, and underwater image understanding. 
However, underwater salient instance segmentation is still a challenging task due to the absence of a large-scale underwater salient instance segmentation dataset for deep learning based methods to effectively train their models. Moreover, underwater images often suffer from unavoidable quality degradation due to wavelength attenuation and scattering. Sea snow formed by plankton in the ocean also introduces noise to the image, which impairs the imaging quality \cite{Akkaynak_2017_CVPR}.
In addition, the specific marine species, water patterns, and large number of optical artifacts in underwater scenes make the visual content of underwater images quite different from that of land-based images \cite{Akkaynak_2018_CVPR}.

Directly transferring conventional SIS methods for land images to underwater scenes may struggle to achieve ideal performance attribute to the domain gap of intrinsic characteristics and extrinsic circumstances between land and underwater living.
They may lose model performance due to low-quality underwater imagery and unique underwater semantic features.
As a result, even state-of-the-art SIS models trained on large-scale land-based datasets are not directly applicable to underwater imagery.
For example in \Cref{fig:simplt_show}, the performance of state-of-the-art methods OQTR \cite{OQTR_2022_TOM}~and RDPNet \cite{RDPnet_2021_TIP}~are degraded on \mAP, \AP{50}, and \AP{75} when applied in underwater. Moreover, we try to combine the SIS method with enhancement pre-processing. Nevertheless, this approach does not seem to work well since it is hard to jointly train multiple tasks to achieve optimal performance.
However, as shown in \Cref{fig:simplt_show} and \Cref{tab:usis.comp},  the performance of such approach is not even able to exceed that of WaterMask \cite{Lian_2023_ICCV}, an underwater instance segmentation method that is not specifically designed for saliency but trained end-to-end.
Establishing a large-scale dataset for the USIS task and designing an underwater effective architecture is necessary.

To address these issues, we propose the first large-scale Underwater Image Salient Instance Segmentation dataset (\datasetName) to explore the application of the SIS field in underwater scenes.
\datasetName~contains 10,632 images and manually labeled pixel-level masks for salient instances in the images.
While previous SIS datasets were usually not labeled with category labels for salient instances, research in the field of weakly supervised salient instance segmentation \cite{tian_2022_IJCV}~has demonstrated that category labeling helps the network to detect semantically dominant regions, offering supplementary information for SIS task training.
Therefore, we labeled each salient instance with a distinct category, including fish, coral reefs, underwater plants, human divers, robots, underwater ruins, and seafloor reefs. Some samples of this dataset are shown in \Cref{fig:usis10k_show}.
Moreover, we propose an Underwater Salient Instance Segmentation model based on SAM, called \methodName.
Since it has been proven that SAM has limited performance in certain tasks due to the lack of domain-specific knowledge \cite{chen2023sam, chen2023rsprompter, wu2023medical}, 
inspired by prompt learning, we devise an Underwater Adaptive
Visual Transformer (UA-ViT) encoder in the proposed \methodName~ to adjust image features as soft prompts, aiming to facilitate the model adapt to the complex underwater scenes. 
In addition, SAM relies on user-supplied point or bounding box coordinates to guide network segmentation.
Therefore, we also design an underwater Salient Feature
Prompter Generator (SFPG) that can directly generate saliency prompts in an image as input to SAM mask decoder for end-to-end segmentation of salient instances in underwater images.

We have performed extensive experiments to evaluate the effectiveness of our method.
First, we compared \methodName~with mainstream salient instance segmentation methods \cite{Fan_2019_CVPR, RDPnet_2021_TIP, OQTR_2022_TOM}~and instance segmentation methods \cite{SAM_2023_ICCV, Lian_2023_ICCV, chen2023rsprompter}~on the \datasetName~to validate the effectiveness of \methodName.
As the simple case shown in \Cref{fig:simplt_show}  our \methodName~can segment more accurate masks and achieve the best quantitative performance.
In addition, we also performed comparisons on the largest extant land salient instance segmentation dataset, SIS10K \cite{OQTR_2022_TOM}, to validate the generalization ability of our model.
The main contributions are concluded as follows:
\vspace{-2mm}
\begin{itemize}
\item We construct the first large-scale dataset, \datasetName, for the underwater salient instance segmentation task, which contains 10,632 images and pixel-level annotations of 7 categories. As far as we know, this is the largest salient instance segmentation dataset, and includes Class-Agnostic and Multi-Class labels simultaneously.
\item We first attempt to apply SAM to underwater salient instance segmentation and propose \methodName, aiming to improve the segmentation accuracy in complex underwater scenes.
We design Underwater Adaptive ViT Encoder to incorporate underwater visual prompts into the network via adapters, 
and Salient Feature Prompter Generator to automatically generate
salient prompters, guiding an end-to-end segmentation network.
\item Extensive public evaluation criteria and large numbers of experiments verify the effectiveness of our \datasetName~dataset and \methodName. 
\end{itemize}


\section{Related Work}
\label{sec:RelatedWork}
\subsection{Segment Anything Model (SAM)}
\label{subsec:SAMRelatedWork}

SAM \cite{SAM_2023_ICCV}~is a promptable model trained on a large dataset, resulting in a powerful zero-shot generalization. 
Currently, extension work based on SAM explores its capability boundaries, for example, image inpainting \cite{yu2023inpaint}, medical image analysis \cite{ma2024segment, zhang2022segment}, few-shot learning \cite{zhang2023personalize, shi2023generalist}, remote sensing segmentation \cite{zhang2024uv, chen2023rsprompter}, image editing \cite{xie2023edit} and style transfer \cite{liu2023any}.
In normal scenes, SAM has excellent generalization capabilities. However, some research results show that SAM already has some limitations in special scenarios (\eg, low-contrast) and specific tasks (\eg, camouflage instance segmentation) \cite{chen2023sam}.
Similarly, the complex background interference in marine environments and the widespread quality degradation of underwater images also present significant challenges to the segmentation capabilities of SAM.

\subsection{Salient Instance Segmentation (SIS)}
\label{subsec:SISRelatedWork}
SIS is to separate salient objects while distinguishing different salient instances.
Most of the existing SIS methods and datasets are designed for land scenes. 
ILOS dataset \cite{Li_2017_CVPR}~is the first SIS dataset, which contains two thousand images.
S4Net \cite{Fan_2019_CVPR}~migrated the Mask R-CNN architecture \cite{MASK_RCNN_2017}~in instance segmentation to the SIS task,  employing the ROIMasking module to exploit the inherent features of the bounding box. 
Similarly, RDPNet \cite{RDPnet_2021_TIP}~augments all feature layers in the feature pyramid to perform multilevel RoIAlign for better mask prediction.
The SOC dataset \cite{SOC_2018_ECCV}~attempts to eliminate the data bias of the saliency dataset, which contains 3,000 images with saliency instances and 3,000 images without it.
OQTR \cite{OQTR_2022_TOM}~introduces the transformer structure to the SIS domain, which locates salient instances by query instead of hand-designed anchors.
Moreover, they provide a new dataset, SIS10K, containing 10,300 terrestrial images to train various networks.

\begin{figure*}[!t]
  \centering
  \includegraphics[width=0.95\linewidth]{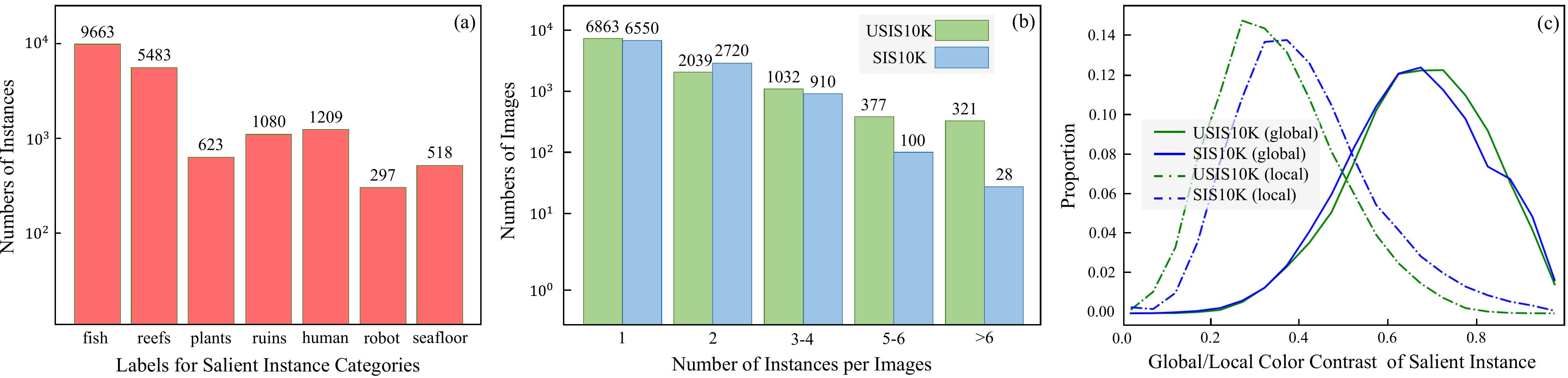}
  \vspace{-4mm}
  \caption{Essential characteristics of the \datasetName~dataset. (a) The number of salient instances per category in the \datasetName~dataset. (b) Distribution of the number of salient instances per image in the \datasetName~dataset. (c) Comparison of \datasetName~and SIS10K in global color contrast and local color contrast.}
  \label{fig:usis10k_statistic_show}
  \vspace{-2mm}
\end{figure*}


\section{\datasetName~Dataset}
\label{sec:Dataset}


\subsection{Dataset Collection and Annotation}
\label{subsec:datasetone}

We have carefully selected 16,000 underwater images from many different natural underwater environments from the Internet and open source underwater datasets \cite{Chongyi_2020_TIP, Dana_2021_PAMI, Lian_2023_ICCV, Hong_2023_TIP, uf0_120_2020}. These images cover underwater vision tasks such as ocean exploration, human-robot intelligent cooperation and underwater autopilot.
Then, the salient instances in these images are labeled pixel by pixel and assigned category labels. To ensure the accuracy and objectivity of the labeling as much as possible, each image will be labeled by at least 3 volunteers, and then the final labeling results will come out through voting.
After removing images that did not have a consensus in the voting or were unclear the instance category, we ended up with a total of 10,632 images for the USIS10K dataset. 
For more in-depth information regarding the collection and annotation of the dataset, please refer to \Cref{app:dataset}.

\begin{table}[!t]
    \vskip -0.1in
    \caption{Comparison with existing SIS datasets. Where Label indicates whether each instance has a separate category label,  Max indicates the maximum number of instances in a single image.}
    \vskip 0.05in
    \begin{center}
    \begin{small}
    \renewcommand{\arraystretch}{1.1}
    \setlength{\tabcolsep}{2.2mm}
    {\begin{tabular}{c|cccccc}
    \hline\hline
    Dataset & Year & Task & Label & Number & Max\\
    \hline
    ILSO & 2017& SIS & $\times$ & 2,000 & 8\\
    SOC & 2018& SIS & \checkmark & 3,000 & 8\\
    SIS10K & 2023 & SIS & $\times$ & \textcolor{blue}{10,300} & 9\\
    USIS10K& 2024 & USIS & \checkmark & \textcolor{red}{10,632} & 9 \\\hline\hline
    \end{tabular}}
    \end{small}
    \end{center}
    \label{tab:sis.comp}
    \vspace{-7mm}
\end{table}

\subsection{Dataset Characteristics and Statistics}
\label{subsec:datasettwo}
In this subsection, we illustrate the basic information, characteristics, and challenges of the \datasetName~dataset. Some statistical information can also be seen in \Cref{fig:usis10k_statistic_show}.

\noindent\textbf{Number and Category of Dataset.}
A large number of underwater images of different scenes, containing various categories, is essential to improve the generalization of the network in complex marine environments and to avoid overfitting. 
For this purpose, we collected 10,632 images from 7 categories and divided the dataset into training, validation, and test sets in the ratio of 7:1.5:1.5. 
As can be seen from \Cref{tab:sis.comp}, \datasetName~is the first salient instance segmentation dataset for underwater scenes.
In addition, in order to make the \datasetName~dataset applicable to more downstream tasks, all the annotations are divided into three parts: category labels, salient instance masks, and bounding boxes. This provides the possibility of using the \datasetName~dataset for weakly supervised salient instance segmentation \cite{tian_2022_IJCV}, salient semantic segmentation \cite{fan_2018_ECCV}, and other downstream tasks in underwater scenes.

\begin{figure}[!t]
  \centering
  \includegraphics[width=0.8\linewidth]{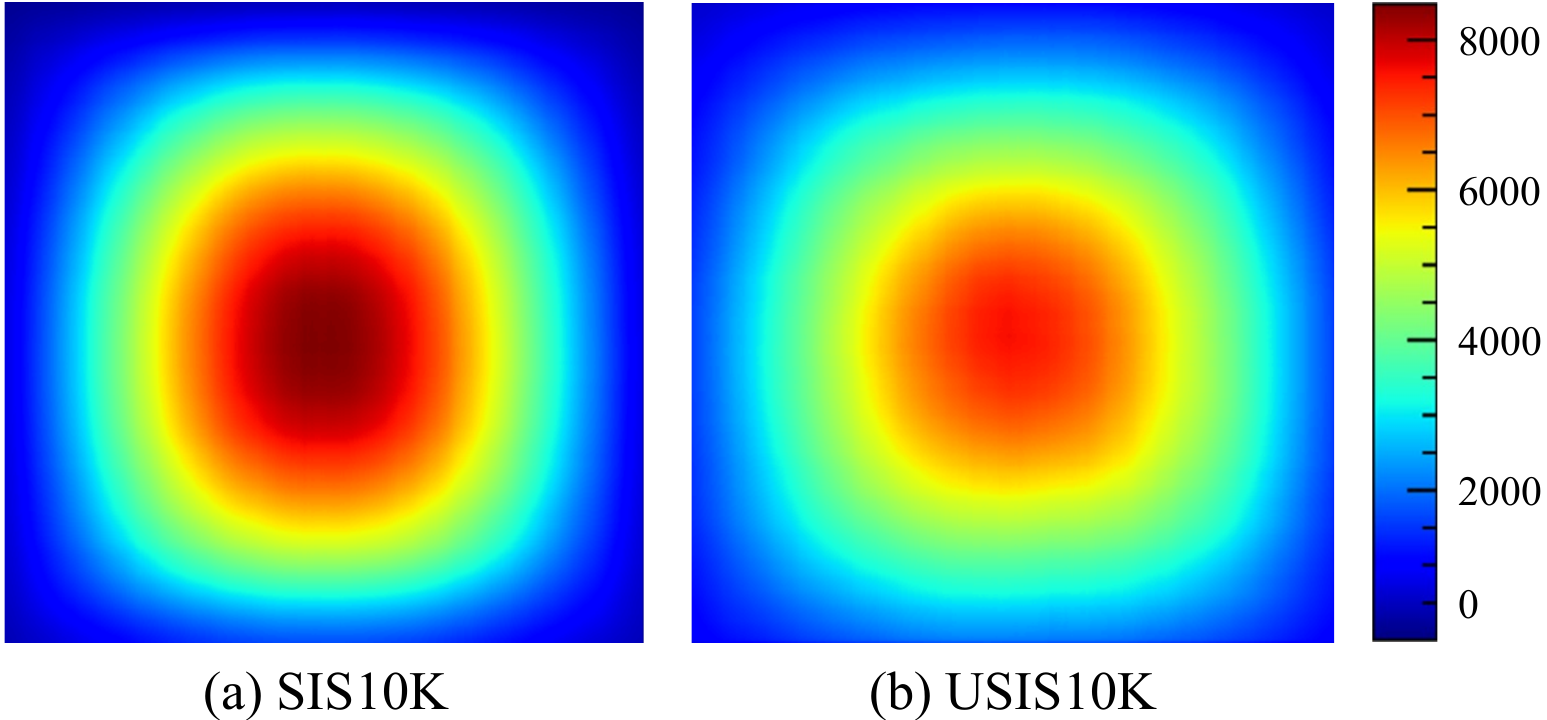}
  \vspace{-4mm}
  \caption{A set of salient maps from our dataset and SIS10K.}
  \vspace{-6mm}
  \label{fig:distribution}
\end{figure}

\noindent\textbf{Number and Size of Salient Instances.}
In \datasetName, multiple salient instances may exist in a single image.
There are 1731 images with more than 3 salient instances in our dataset, accounting for 16.3\% of the total. 
In SIS10K \cite{OQTR_2022_TOM}, there are only 1038 images with more than 3 salient instances, accounting for 10.1\% of the total. 
In the underwater salient detection dataset, USOD10K \cite{Hong_2023_TIP}, there are only 722 images with more than 3 salient objects, accounting for 7.1\% of the total.
This not only means that our \datasetName~is challenging even for state-of-the-art methods, but also allows it to be used in fields like underwater co-salient instance segmentation \cite{cong2022global, zhang2020coadnet, Cosalient_2018_AAAI}.
Furthermore, predicting instances that are too small or too large is a very common but challenging problem.
In the \datasetName~dataset, the average size of the salient instances is 34,336 pixels (approximately 185$\times$185 pixels), which averaged 10.3\% of the image size.
There are 3053 salient instances smaller than 1\% of the image area, (16.0\% of the total), while there are 1733 instances larger than 30\% of the image area, (9.1\% of the total), which further enhances the challenge of \datasetName.

\noindent\textbf{Color Contrast of Salient Instance.}
Saliency is often related to the global contrast between foreground and background, and it is critical to check whether salient instances are easy to detect \cite{li_2014_secrets}. 
We calculated the Bhattacharyya distance \cite{Bhattacharyya_2003_PR}~between the RGB histogram of each salient instance and the background RGB histogram in \datasetName~and SIS10K dataset to measure their global contrast. 
It can be seen that the global contrast of \datasetName~is slightly higher than SIS10K.
In addition, similar to the global contrast, we also computed the local contrast in a 5$\times$5 patch on the salient instance boundary. 
It can be seen that \datasetName~has less local contrast than SIS10K due to characteristics like instance clustering in underwater environments \cite{Lian_2023_ICCV, Hong_2023_TIP}. 
This poses a greater challenge for accurate segmentation of salient instance masks in the boundary portion of the network.

\begin{figure}[t]
  \centering
  \includegraphics[width=0.8\linewidth]{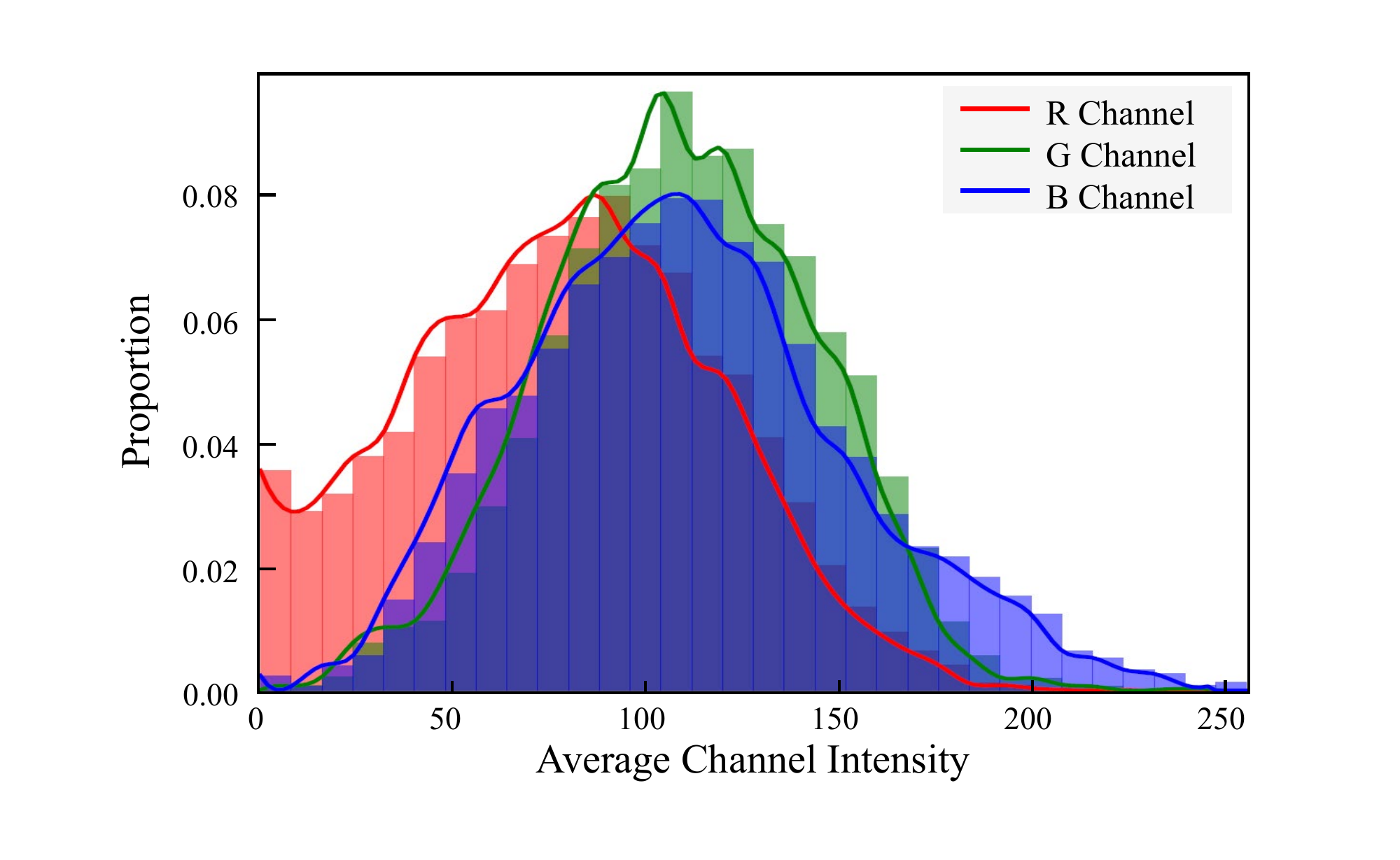}
  \vspace{-4mm}
  \caption{Average channel intensity in \datasetName~with proportion.}
  \vspace{-4mm}
  \label{fig:channelintensity}
\end{figure}

\begin{figure*}[t]
  \centering
  \includegraphics[width=1\linewidth]{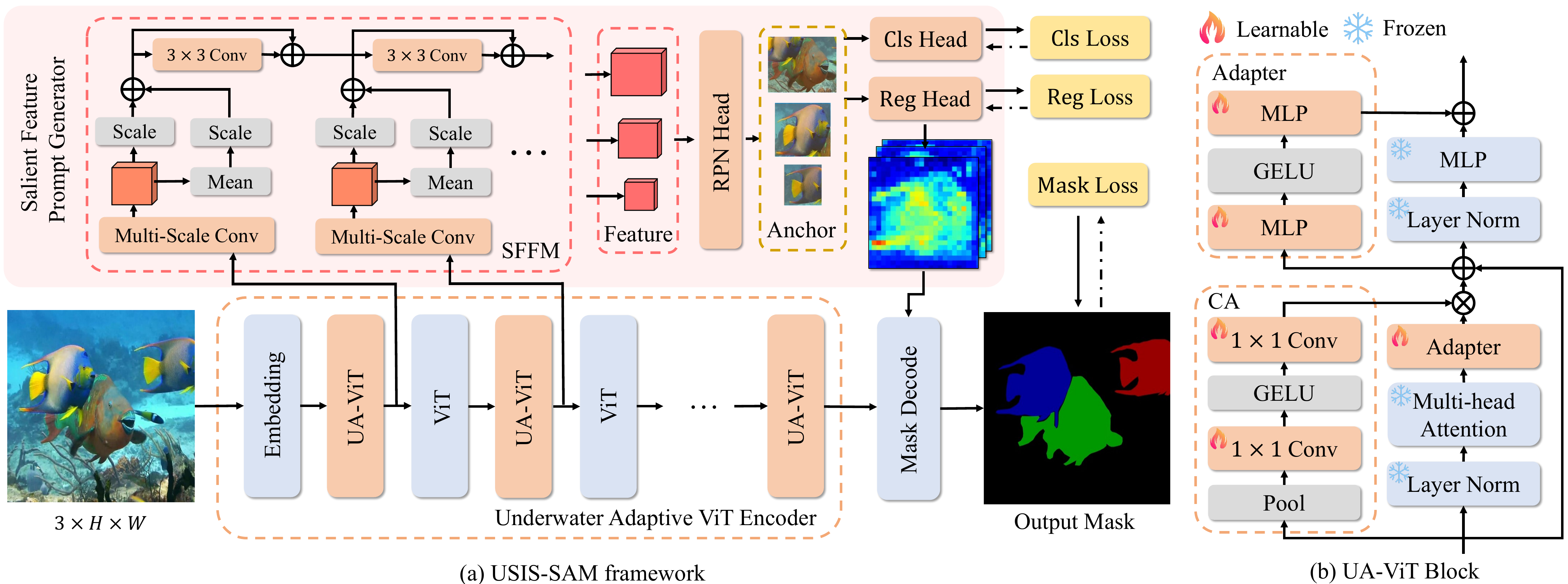}
  \vspace{-8mm}
  \caption{(a) \methodName~framework.~The \methodName~framework modifies the SAM by adding the Underwater Adaptive ViT Encoder (in \Cref{sec:adaptiveEncoder}) and the Salient Feature Prompt Generator (in \Cref{sec:promptGenerator}). (b) The structure of UA-ViT. In the figure, SFFM stands for Salient Feature Fusion Module, CA stands for Channel Adapter.}
  \vspace{-2mm}
  \label{fig:USIS-SAM}
\end{figure*}

\noindent\textbf{Location of Salient Objects.}
The central bias has been identified as one of the most significant biases in salience datasets \cite{SOC_2018_ECCV, Hong_2023_TIP}~due to the fact that humans have a natural tendency to perceive salient objects by focusing their attention on the center of the observed scene. 
\Cref{fig:distribution}~illustrates a set of salient instance heat maps. It can be seen that while \datasetName~has a smaller central deviation compared to SIS10K. 
Specifically, in the SIS10k dataset, approximately 13.5\% of the locations have fewer than 1,000 instances and 32\% have fewer than 2,000 instances, while in our dataset, only 2.75\% of the locations have fewer than 1,000 instances and 22.5\% have fewer than 2,000 instances.

\noindent\textbf{Channel Intensity of Underwater Images.}
Optical images inevitably suffer from color attenuation due to the selective absorption of water at different wavelengths.
In particular, the attenuation of the red channel is an order of magnitude greater than that of the blue and green channels \cite{mobley2001radiative}.
As a result, in most cases, underwater images are bluish or greenish. 
To quantify this attenuation property of the \datasetName, we compute the average channel intensities and probability densities of R, G, and B for each image in \Cref{fig:channelintensity}.
It can be seen that R channel has the lowest intensity, but also follows the same trend as G and B channels.

\section{\methodName}
In this section, we introduce our proposed \methodName, a prompt learning method based on SAM. The framework of it is illustrated in \Cref{fig:USIS-SAM} (a).
In USIS-SAM, we design the Underwater Adaptive ViT (UA-ViT) to integrate underwater visual prompts into the network via adapter and channel adapter. A Salient Feature Prompt Generator (SFPG) was also designed to aggregate the UA-ViT output features and generate salient prompts to guide end-to-end segmentation.

\subsection{Underwater Adaptive ViT Encoder}
\label{sec:adaptiveEncoder}

Due to being trained on the SA-1B dataset \cite{SAM_2023_ICCV}, the largest existing natural image segmentation dataset, the SAM image encoder possesses strong capabilities in extracting image features. 
This makes it a valuable tool for underwater image salient segmentation. 
However, underwater images often suffer from color distortion, and the SAM image encoder lacks expertise in the underwater domain due to the domain bias between the SA-1B dataset and the complex marine environment.
Consequently, the SAM image encoder is unable to fully extract features from underwater images, and its direct application to underwater scenes does not yield satisfactory results.
To address this issue, we propose the Underwater Adaptive ViT (UA-ViT) inspired by the Parameter-Efficient Fine-Tuning (PEFT) technique \cite{bitfit_2022_acl}.
By incorporating a set of adapters into ViT block (as depicted in \Cref{fig:USIS-SAM} (b)), the UA-ViT enables a more effective utilization of the SAM image encoder in underwater scenarios.


UA-ViT learns the expertise of the underwater visual prompts via adapters and injects it into the subsequent network. 
Essentially, the adapter can be regarded as a soft prompt to embed the network, which takes advantage of the fact that the ViT has been trained on large-scale datasets, selectively mask or leverage the original network structure and weights by adjusting the inflow features, and enhances the model's ability to generalize to downstream tasks with only a small amount of training.
Therefore, we added two adapters for each ViT block. 
Specifically, we place the first adapter immediately after multi-head attention layer to adjust for color distortion due to selective absorption of water at different wavelengths.
The second adapter on the residual path of the frozen MLP layer, and utilizes parallel branches to aggregate the updated context by summation while maintaining the original MLP features. 
Where the adapter can be represented as:
\vspace{-1.5mm}
\begin{equation}
  P=MLP_{out}(\sigma (MLP_{prompt}(F))),
  \label{eq:adapter}
  \vspace{-1.5mm}
\end{equation}
where $MLP_{prompt}$~is the linear layer used to generate the underwater task prompts for each adapter, and $MLP_{out}$~is the projection layer used to resize the adaptive features. 
$F$~is the input feature, and $P$~is the output prompt for each adapter layer. 
$\sigma$~is the activation function. 

We further propose a channel adapter (CA) module to address the frequency distribution bias due to the selective absorption of wavelengths by the water, which utilizes global attention-aware features to adaptively adjust the importance of each channel, thus significantly improving the representation capability of the network.
The channel adapter can be represented as:
\vspace{-1mm}
\begin{equation}
  C=F*Conv_{up}(\sigma (Conv_{down}(Pool(F)))),
  \label{eq:channel_sa}
  \vspace{-1mm}
\end{equation}
where $C$~is the output feature after channel adapter, $Conv$~ is a 1$\times$1~convolutional layer, and $Pool$~is an average pooling layer.
It can be seen that the channel adapter has a very similar structure to the adapter, only the MLP is replaced by the convolution.
In fact, they share a similar function, and the channel adapter focuses more on the adjustment of frequency domain information to help the network adapt to the underwater environment.

In order to minimize the number of parameters, we replace only some of the ViT blocks in the SAM encoder with UA-ViT blocks.
Specifically, when we use ViT-H \cite{ViT_2021_ICLR}~as the backbone of SAM encoder, we replace one layer every two layers starting from the eighth layer.


\subsection{Salient Feature Prompt Generator (SFPG)}
\label{sec:promptGenerator}
The USIS task needs the model to automatically recognize and segment each salient object in underwater images.
However, SAM requires the user to explicitly provide foreground points, boxes, or texts as prompts to guide the model segmentation. 
Simply adding the object detection network to identify the location of significant objects and feeding the bounding boxes as prompts into the SAM's prompt encoder would significantly increase the complexity of the model and severely limit the model optimization. Therefore, we design the Salient Feature Prompt Generator to directly predict prompts embedding of salient instances.

The SFPG first designs a Salient Feature Fusion Module (SFFM) to fuse the features output from each UA-ViT block and align them with the prompts required by the SAM decoder.
We first aggregate multi-scale features through multi-scale convolutions, suppress irrelevant features, and reasonably capture significant information in images. The multi-scale features of the ith-layer UA-ViT $F^m_i$~can be obtained by:
\vspace{-1mm}
\begin{equation}
  F^m_i = \sum_{s\in {3,5,7}} Conv_{s*s}(CA(F_i)),
  \label{eq:fi}
  \vspace{-1mm}
\end{equation}
where $F_i$ is the features from the ith-layer UA-ViT, $Conv_{s*s}$ are the multi-scale convolutions ($3 \times  3$, $5 \times  5$, and $7 \times  7$), and $CA$ means channel adapter. 
We then balance the multi-scale features using the average residuals to dampen the noise in the features, which can be formulated as:
\vspace{-1mm}
\begin{equation}
  F^a_i = \lambda F^m_i + (1-\lambda)Avg(F^m_i),
  \label{eq:mean}
  \vspace{-0.5mm}
\end{equation}
where $Avg$~is the global average pooling, $\lambda$~is a hyperparameter that controls the noise suppression effect, which in this paper is 0.8.
In addition, the ith $F^a_i$~will also be fused with the previous feature, which can be denoted as:
\vspace{-1mm}
\begin{equation}
  F^{a*}_i = F^a_i  + Conv_{3*3}(F^{a*}_{i-1}).
  \label{eq:add}
  \vspace{-0.5mm}
\end{equation}
Finally, since the saliency of the instance varies with the image scale, we use multi-scale deconvolution to 2$\times$~and 4$\times$~upsampling of the final output $F^{a*}_{n}$~of the SFFM, which is used in the subsequent RPN header \cite{FasterRCNN_2015_NIPS}.

\begin{figure}[!t]
  \centering
  \includegraphics[width=1\linewidth]{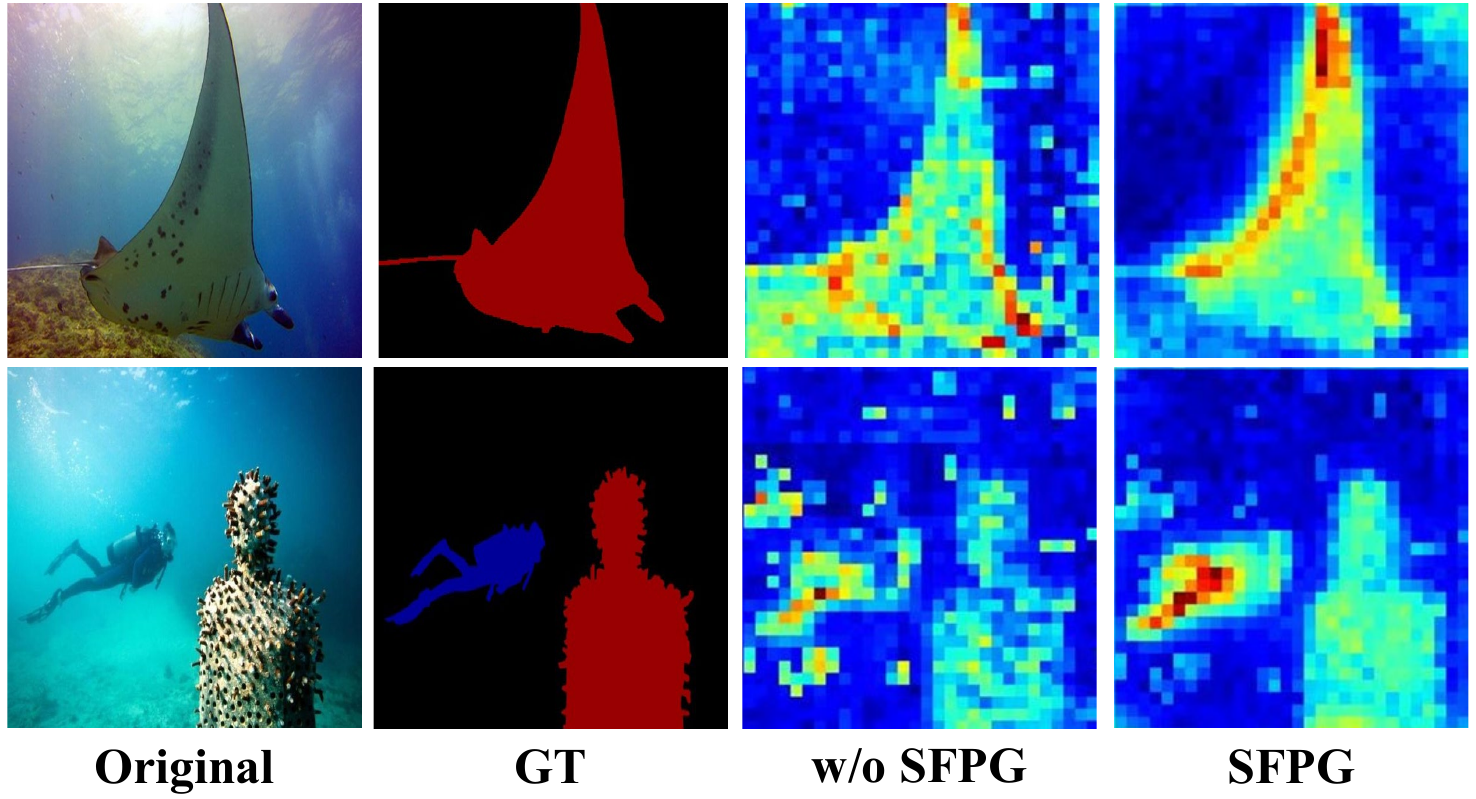}
  \vspace{-8mm}
  \caption{Visualize features generated by the SFPG. The SFPG can aggregate features on salient instances.}
  \vspace{-2mm}
  \label{fig:salient_attention}
\end{figure}

As a visual illustration in \Cref{fig:salient_attention}, we can clearly observe a significant change in the distribution of salient features with application SFPG module.
This is due to the ability provided by the SFPG module to aggregate different receptive field features, where large receptive field features help the model to locate salient regions while filtering out noise, and small receptive field features contribute to supplementing the image details with fine-grained semantic granularity.

\noindent\textbf{Loss Function.} 
The loss function of the USIS-SAM framework is similar to the Mask RCNN \cite{MASK_RCNN_2017}~and consists of several components \ie, localization loss, classification loss, regression loss and segmentation loss. Therefore, the cumulative loss can be expressed as:
\vspace{-1mm}
\begin{equation}
  \mathcal L= {\frac{{\sum_{i}^{N} \mathcal L^i_{rpn}}}{N}}  + {\frac{{\sum_{i}^{N} \mathbbm{1}(i) \cdot (\mathcal L^i_{cls} + \mathcal L^i_{reg} + \mathcal L^i_{seg})}}{\sum^N_i \mathbbm{1}(i)}} ,
  \label{eq:loss}
  \vspace{-0.25mm}
\end{equation}
where $N$ is the total number of anchor boxes predicted by the RPN head and $\mathbbm{1}(\cdot)$ is the indicator function, which outputs 1 when the anchor box is salient and otherwise outputs 0. $\mathcal L^i_{rpn}$ denotes the region proposal loss, with the same configuration as Mask RCNN \cite{MASK_RCNN_2017}. $\mathcal L^i_{cls}$ denotes the cross-entropy loss, computed between the predicted categories and the ground truth categories. $\mathcal L^i_{reg}$ is the Smooth L1 loss \cite{girshick2015fast}, computed based on the bounding box coordinate offset between the prediction and the ground truth. $\mathcal L^i_{seg}$ is the binary cross-entropy loss between the USIS-SAM output salient mask and the ground truth mask.


\begin{table*}[ht]
    \vspace{-4mm}
    \caption{Quantitative comparisons with state-of-the-arts on the \datasetName~datasets. URank stands for the underwater image enhancement method in UnderwaterRanker \cite{underwaterranker_aaai_2023}. 
    SAM+BBox uses inference results from Faster RCNN \cite{FasterRCNN_2015_NIPS}~as prompts for prediction, SAM+Mask stands for Mask RCNN networks \cite{MASK_RCNN_2017}~use SAM as the backbone.
    The RSPrompter in the table is the RSPrompter-anchor framework. }
    \vspace{-3mm}
    \begin{center}
    \renewcommand{\arraystretch}{1.1}
    \setlength{\tabcolsep}{2.5mm}
    {\begin{tabular}{c|c|c|ccc|ccc}
    \hline\hline
    \multirow{2}{*}{Method} & \multirow{2}{*}{Epoch} & \multirow{2}{*}{Backbone}  &\multicolumn{3}{c|}{Class-Agnostic} & \multicolumn{3}{c}{Multi-Class}\\ \cline{4-9}
     &  &   &  \mAP & \AP{50} & \AP{75} & \mAP & \AP{50} & \AP{75}\\
    \hline
    S4Net \cite{Fan_2019_CVPR} & 60 & ResNet-50 &  32.8 & 64.1 & 27.3 & 23.9 & 43.5 & 24.4 \\
    RDPNet \cite{RDPnet_2021_TIP} & 50 & ResNet-50 & 53.8 & 77.8 & 61.9 & 37.9 & 55.3 & 42.7\\
    RDPNet \cite{RDPnet_2021_TIP} & 50 & ResNet-101  & 54.7 & 78.3 & 63.0 & 39.3 & 55.9 & 45.4\\
    OQTR \cite{OQTR_2022_TOM} & 120 & ResNet-50 &  56.6 & 79.3 & 62.6 & 19.7 & 30.6 & 21.9\\
    URank+RDPNet \cite{RDPnet_2021_TIP} & 50 & ResNet-101  & 52.0 & 80.7 & 62.0 & 35.9 & 52.5 & 41.4\\
    URank+OQTR \cite{OQTR_2022_TOM} & 120 & ResNet-50 &  49.3 & 74.3 & 56.2 & 20.8 & 32.1 & 23.3\\
    \hline
    WaterMask \cite{Lian_2023_ICCV} & 36 & ResNet-50  & 58.3 & 80.2 & 66.5 & 37.7 & 54.0 & 42.5 \\
    WaterMask \cite{Lian_2023_ICCV} & 36 & ResNet-101  & 59.0 & 80.6 & 67.2 & 38.7 & 54.9 & 43.2\\
    SAM+BBox \cite{SAM_2023_ICCV} & 24 & ViT-H  & 45.9 & 65.9 & 52.1 & 26.4 & 38.9 & 29.0 \\
    SAM+Mask \cite{SAM_2023_ICCV} & 24 & ViT-H  & 55.1 & 80.2 & 62.8 & 38.5 & 56.3 & 44.0 \\
    RSPrompter \cite{chen2023rsprompter} & 24 & ViT-H  & 58.2 & 79.9 & 65.9 & 40.2 & 55.3 & 44.8\\
    URank+RSPrompter \cite{chen2023rsprompter} & 24 & ViT-H  & 50.6 & 74.4 & 56.6 & 38.7 & 55.4 & 43.6\\
    \hline
    \methodName & 24 & ViT-H  & \textbf{59.7} & \textbf{81.6} & \textbf{67.7} & \textbf{43.1} & \textbf{59.0} & \textbf{48.5}\\
    \hline  \hline
    \end{tabular}}
    \end{center}
    \label{tab:usis.comp}
    \vspace{-5mm}
\end{table*}

\section{Experiments}

We evaluated our \methodName~compared with other state-of-the-art methods on \datasetName.
First, since previous SIS datasets are usually class-agnostic, the experiments only require the network to segment each salient instance in the image without predicting their classes. 
We followed previous conventions in the SIS field and performed class-agnostic salient instance segmentation.
Then, we also require the network to perform multi-class salient instance segmentation, \ie, predicting instance labels while segmenting salient instances, in order to evaluate the overall capability of the network.
We follow OQTR \cite{OQTR_2022_TOM}~and adopt the standard mask AP metrics \cite{mscoco_2014_ECCV}~as evaluation Criteria, including \mAP, \AP{50}~and \AP{75}.

\subsection{Implementation Details}

In this paper, all backbones and hyperparameters of the methods and comparison algorithms are the same as original paper, except for our newly designed parts. 
We implemented USIS-SAM with PyTorch and MMDetection \cite{mmdetection}, and trained it over 24 epochs on 6 NVIDIA 3090 GPUs, using the AdamW optimizer with a starting learning rate of 1e-4 and weight decay of 1e-3. We also used the MindSpore Lite tool\footnote{\url{https://www.mindspore.cn/}} to implement our network.
Unless otherwise stated, the hyperparameters of all ablation experiments were consistent with the comparison experiments with multi-class salient instance segmentation.
For data enhancement, both our and the comparison algorithms use random horizontal flipping, random scaling, and random cropping.
During training, the original parameters of SAM module were frozen for our USIS-SAM and other compared methods.

\subsection{Experimental Results}

\begin{figure*}[t]
  \centering
  \includegraphics[width=1\linewidth]{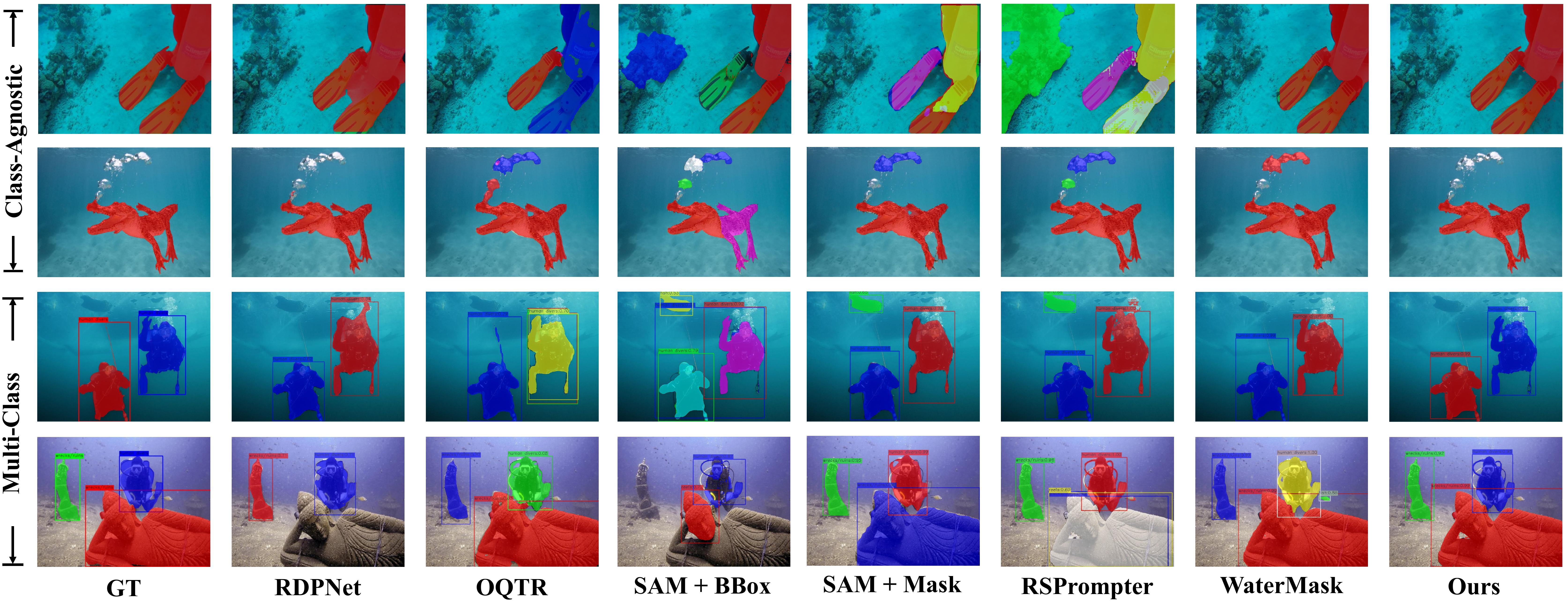}
  \vspace{-8mm}
  \caption{Qualitative comparison on the USIS10K dataset. Each salient instance is represented by a unique color, and the segmented mask is superimposed on the image. More quantitative comparisons can be found in \Cref{fig:supp_qc} in the Appendix.}
  \vspace{-2mm}
  \label{fig:qualitative}
\end{figure*}

In this subsection, we will test class-agnostic salient instance segmentation and multi-class salient instance segmentation for \methodName~and other state-of-the-art methods.
The relevant results of the quantitative analysis and qualitative comparison can be found in \Cref{tab:usis.comp} and \Cref{fig:qualitative}. More experiments can be found in \Cref{app:more_experiments}.

\noindent\textbf{Class-Agnostic Salient Instance Segmentation.} 
Class-agnostic salient instance segmentation can be essentially understood as foreground instance segmentation focused on salient parts in images.
Thus, the model needs a robust capability to localize salient regions, and the segmented salient instance masks must strictly align with the ground truth to achieve good experimental results in this task.
It can be seen that \methodName~outperforms the best-performing SIS method, OQTR, by 3.1, 2.3, and 5.1 AP in \mAP, \AP{50}, and \AP{75}.
This indicates that the AP advantage of \methodName~is due to the ability to segment higher precision masks in the underwater environment.
In contrast, \methodName~has an advantage of 0.7, 1.0, and 0.5 AP in \mAP, \AP{50}, and \AP{75} compared to the underwater instance segmentation method WaterMask.
At this point, the AP advantage of \methodName~is based on the accurate identification of salient areas.
Regarding the RSPrompter, also built based on SAM, \methodName~outperforms it by 1.5, 1.7, and 1.8 AP on \mAP, \AP{50}, and \AP{75}.
This improvement is attributed to better learning of the underwater visual prompt with the assistance of the Underwater Adaptive ViT Encoder in \methodName, enhancing the network's accuracy in segmentation.


\noindent\textbf{Multi-Class Salient Instance Segmentation.} 
Multi-class salient instance segmentation can be considered as a fusion of tasks from Class-Agnostic Salient Instance Segmentation and Salient Instance Class Prediction.
From \Cref{tab:usis.comp} in the Multi-class case, \methodName~continues to gain an AP advantage due to better mask accuracy. 
For OQTR, S4Net, the single-stage network coupling classification and segmentation, \methodName~lead them by 22.4 AP and 19.5 AP.
Furthermore, \methodName~also gains a 4.1, 3.3, and 4.0 AP lead over two-stage network RDPNet for \mAP, \AP{50}, and \AP{75}.
Moreover, \methodName~achieved a 4.4 and 5.1 AP lead in \mAP~over WaterMask and RSPrompter, which were originally designed for multi-class instance segmentation tasks. 
\methodName~leads them by 5.1 and 6.2 AP on \AP{50}, respectively.
This is benefit from the extra labels that help the network localize semantically dominant regions.

\begin{table}[!t]
    \caption{Effectiveness of each component in \methodName, replacing SFPG means to use Multi-scale Feature Enhancer Module \cite{chen2023rsprompter} instead of SFPG.}
    \begin{center}
    \renewcommand{\arraystretch}{1.1}
    {\begin{tabular}{c|ccc}
    \hline\hline
    Methods & \mAP & \AP{50} & \AP{75} \\
    \hline
    Full Model & 43.1 & 59.0 & 48.5\\
    w/o UA-ViT & 41.5 (-1.6) & 57.4 (-1.6) & 47.0 (-1.5)\\
    replace SFPG &  42.2 (-0.9) &58.3 (-0.7) &47.5 (-1.0)  \\
    \hline\hline
    \end{tabular}}
    \end{center}
    \label{tab:ab}
    \vspace{-2mm}
\end{table}

\subsection{Ablation Studies}
In this subsection, we perform ablation experiments on Underwater Adaptive ViT Encoder and Salient Feature Prompt Generator, and the relevant data can be found in \Cref{tab:ab}.  We also performed comparisons on the largest
land salient instance segmentation dataset, SIS10K
 \cite{OQTR_2022_TOM}, to validate the generalization ability of our
model.
More detailed ablation experiments are in \Cref{subsec:more_ablation}.

\noindent\textbf{Effectiveness of Underwater Adaptive ViT Encoder.} 
We validate the performance of the Underwater Adaptive ViT Encoder by reverting the UA-ViT Block added in USIS-SAM back to the original ViT Block.
In other words, at this stage, USIS-SAM will utilize the pre-trained ViT-H as the image encoder. It can be observed that the model achieves a 1.6 AP improvement after incorporating the UA-ViT Block. This demonstrates that the UA-ViT Block introduces underwater visual information to the Underwater Adaptive ViT Encoder through adapters, enabling the network to effectively handle complex marine scenes such as marine snow, light scattering, optical artifacts, etc. Consequently, it enhances the network's capability to extract features and facilitates network inference.

\noindent\textbf{Effectiveness of Salient Feature Prompt Generator.} 
We evaluate the effect of the Salient Feature Prompt Generator by replacing it. 
After replacing, the network will utilize the Multi-scale Feature Enhancer \cite{chen2023rsprompter}~to aggregate multi-layer features.
After the replacement, the network's \mAP, \AP{50}, and \AP{75} are reduced by 0.9, 0.7, and 1.0 AP, respectively.
What makes this possible is that SFPG do not simply fuse backbone features, it also enables the network to filter attention from non-significant regions and direct segmentation gravity to salient locations.

\begin{table}[!t]
    \caption{Quantitative comparisons with state-of-the-arts methods on the SIS10K \cite{OQTR_2022_TOM}. The hyperparameters and other settings for the methods in the table are the same as in \Cref{tab:usis.comp}.}
    \vskip 0.05in
    \begin{center}
    \setlength{\tabcolsep}{3.2mm}
    \renewcommand{\arraystretch}{1.1}
    {\begin{tabular}{c|ccc}
    \hline\hline
    Methods & \mAP & \AP{50} & \AP{75} \\
    \hline
    OQTR \cite{OQTR_2022_TOM}& 67.2 & 88.1 & \textbf{81.7}\\
    USIS-SAM & \textbf{70.1} & \textbf{89.0} & 78.2\\
    \hline\hline
    \end{tabular}}
    \end{center}
    \label{tab:sis_C}
    \vspace{-2mm}
\end{table}

\noindent\textbf{Generalization Ability of USIS-SAM.}
Since there is no other underwater instance segmentation dataset and it is also a thought-provoking
idea whether USIS-SAM can be directly applied to land imagery, we retrained USIS-SAM on saliency instance segmentation dataset SIS10K to validate the generalization ability of our model. In \Cref{tab:sis_C}, we can see that, contributing to the excellent segmentation ability of SAM and the saliency region localization ability of SFPG, USIS-SAM still achieves good results compared to OQTR, and gains on AP50, but is slightly weaker on AP75.
This may be due to domain knowledge learned in UA-ViT is the same as SAM encoder learned on SA-1B dataset, and the guiding effect of UA-ViT on the network is weakened. This shows that USIS-SAM did not overfit our dataset. 

\section{Discussion}

USIS-SAM proposed in this work is devised as the baseline of USIS10K dataset for underwater scenes, its prompt learning architecture may be feasible when transferred to other domains. 
For example, dark light and hazing/raining scenes share the similar atmospheric scattering model \cite{akkaynak2019sea, zhao2021refinednet}~with underwater scenes. Hence, UA-ViT's adapter architecture may be conducive to
 them automatically map color distortion features to corresponding enhancement features.
Meanwhile, the Salient Feature Prompt Generator merges multi-layer and multi-scale features to provide salient prompts, can serve as an out-of-the-box module to facilitate other SIS methods to further explore and utilize SAM.

\section{Conclusions}

In this paper, we construct the first large-scale salient instance segmentation dataset for underwater images, \datasetName, with pixel-level annotations, which enables us to explore the application of salient instance segmentation techniques in underwater environments.
In addition, we also propose \methodName~for underwater salient instance segmentation based on SAM. 
Extensive experiments demonstrate the effectiveness of the \datasetName~dataset and validate the performance of \methodName.

\section*{Acknowledgements}

This work was supported in part by the National Key R\&D Program of China 2022ZD0118300; in part by the National Natural Science Foundation of China under Grant 62201179;  in part by the  Innovation Platform for ``New Star of South China Sea" of Hainan Province under Grant No. NHXXRCXM202306; in part by the specific research fund of The Innovation Platform for Academicians of Hainan Province under Grant No.YSPTZX202410; in part by the Research Start-up Fund of Hainan University (No. KYQD(ZR)-22015); in part by the Taishan Scholar Project of Shandong Province under Grant tsqn202306079,  in part by Xiaomi Young Talents Program, in part by the Hong Kong GRF-RGC General Research Fund under Grant 11209819 and Grant 11203820.

\section*{Impact Statement}
This paper presents work whose goal is to advance the field of Machine Learning. There are many potential societal consequences of our work, none which we feel must be specifically highlighted here.

{
    \small
    \bibliographystyle{icml2024}
    \bibliography{usis}
}

\newpage
\appendix
\onecolumn
\section{More Details and Samples about USIS10K}

\label{app:dataset}
%

\subsection{Dataset Collection.}
We collected about 30,000 images from the Internet and open source underwater datasets from various different domains, such as underwater image enhancement \cite{Chongyi_2020_TIP, Dana_2021_PAMI}, underwater instance segmentation \cite{Lian_2023_ICCV}, and underwater salience detection \cite{Hong_2023_TIP, uf0_120_2020}.
These images come from many different natural underwater environments, covering underwater vision tasks such as ocean exploration, human-computer intelligence cooperation, and underwater autopilot. 
After that, we allowed two volunteers to filter the alternative images for duplicates, corruptions, and images without salient objects. Finally, we carefully annotated the remaining 16,000 images.

\subsection{Dataset Annotation.}
We assembled 17 volunteers to annotate the dataset. Before annotation, these volunteers were trained in the following three parts:
1) the categories of various common organisms in underwater scenes. 
2) the process of previous dataset construction in the field of SIS and the basic methods for image annotation. 
3) the general knowledge of visible salience. 
We use sparse annotation polygons to annotate each instance in the dataset. Annotated data will be stored in the mainstream COCO-style format \cite{mscoco_2014_ECCV}~for ease of use by most common frameworks and models.
Each image will be annotated by at least 3 volunteers. 
For those images collected from the Underwater Instance Segmentation dataset, 3 volunteers will vote on the saliency of each instance, and all instances with more than 2 votes will be retained.
For images collected from the underwater salient detection dataset, two volunteers will re-label the salient object bounding box at the instance level, and a third volunteer will select the best annotation and refine it.
For the other images, three volunteers will annotate salient objects in the image at the instance level, and a fourth volunteer will select, refine, and merge their annotations.
To ensure the precision of the labeled categories, we follow the guidance of the literature \cite{conf2001Oceana, conf2009ETI} to categorize potentially confusing objects. In addition, we also filtered out the images for which we could not reach a consensus on instance category or salience.
Finally, we are left with 10,632 images that constitute the \datasetName~dataset

\subsection{Category labels contained in \datasetName}
\label{subsec:more_sample}

\Cref{tbl:label}~lists all the category labels contained in the \datasetName~dataset along with their detailed category definitions.
As can be seen, the USIS10K dataset contains object categories such as fish, coral reefs, aquatic plants, and wrecks/ruins.
These categories constitute the primary research components for marine ruins discovery, marine resources exploration, and marine ecological preservation. 
Moreover, the dataset contains pixel annotations for human divers, robots, and seafloor, serving as crucial research targets for human-robot-object collaboration.

\begin{table}[!ht]
    \vskip -0.1in
    \caption{The category labels in \datasetName~dataset with their definitions.}
    \vskip 0.05in
    \begin{center}
    \scalebox{0.94}{
    \renewcommand{\arraystretch}{1.1}
        \begin{tabular}{c|c}\hline\hline
        \textbf{Category} & \textbf{Descriptions}\\\hline
        Fish & Underwater vertebrates, \eg, fish, turtles \\\hline
        Reefs & Underwater invertebrates and coral reefs \\\hline
        Aquatic plants & Aquatic plants and flora \\\hline
        Wrecks/ruins & Wrecks, ruins and damaged artifacts \\\hline
        Human divers & Human divers and their equipment \\\hline
        Robots &  Underwater robots like AUV, ROV \\\hline
        Sea-floor & Rocks and reefs on the seafloor \\
        \hline\hline\end{tabular}
    }
    \end{center}
    \vskip -0.05in\label{tbl:label}
\end{table}

\subsection{More Samples about \datasetName}
\label{subsec:more_sample_ill}

\begin{figure*}[!t]
  \centering
  \includegraphics[width=1\linewidth]{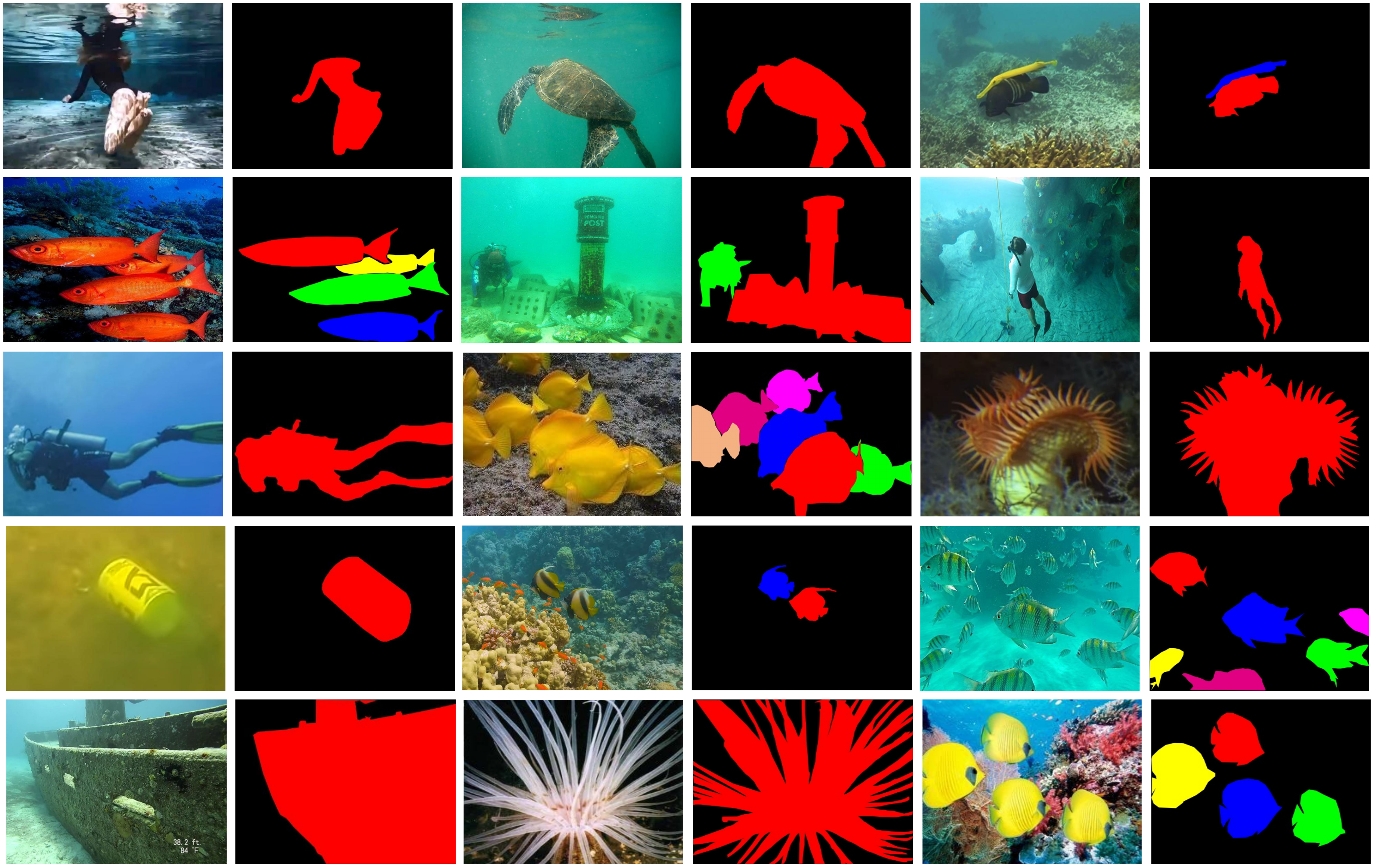}
  \caption{\textbf{More visual samples of annotated images with pixel-level salient instance segmentation in \datasetName.} The image on the left is the original image and the right is the annotation mask, different colors represent different salient instances.}
  \label{fig:supp_usis10k_show}
\end{figure*}

More visual samples from the USIS10K dataset are presented in \Cref{fig:supp_usis10k_show}. 
It is evident that the images in the \datasetName~dataset contain a wide variety of underwater scenes with many different saturations, chromatic aberration, and quality degradations. 
This diversity is very effective for training salient instance segmentation networks suitable for general underwater scenes.

Furthermore, the \datasetName~dataset also includes a large number of challenging images meticulously crafted to comprehensively evaluate the model's performance.
As in \Cref{fig:supp_usis10k_show}, the surface reflection in the first row's, first, and second columns has the potential to disturb the model's saliency localization. 
The intricate boundaries of aquatic plants in the third column of the third row and reefs in the second column of the fourth row present a more significant challenge to the segmentation accuracy of the network.
Additionally, the overlapping fish in the second column of the third row necessitates that the network accurately distinguishes each salient instance.

\section{More Qualitative Comparison and ablation Experiments}
\label{app:more_experiments}

\subsection{More Qualitative Comparison}
\label{subsec:more_qualitative}

\begin{figure*}[!t]
  \centering
  \includegraphics[width=1\linewidth]{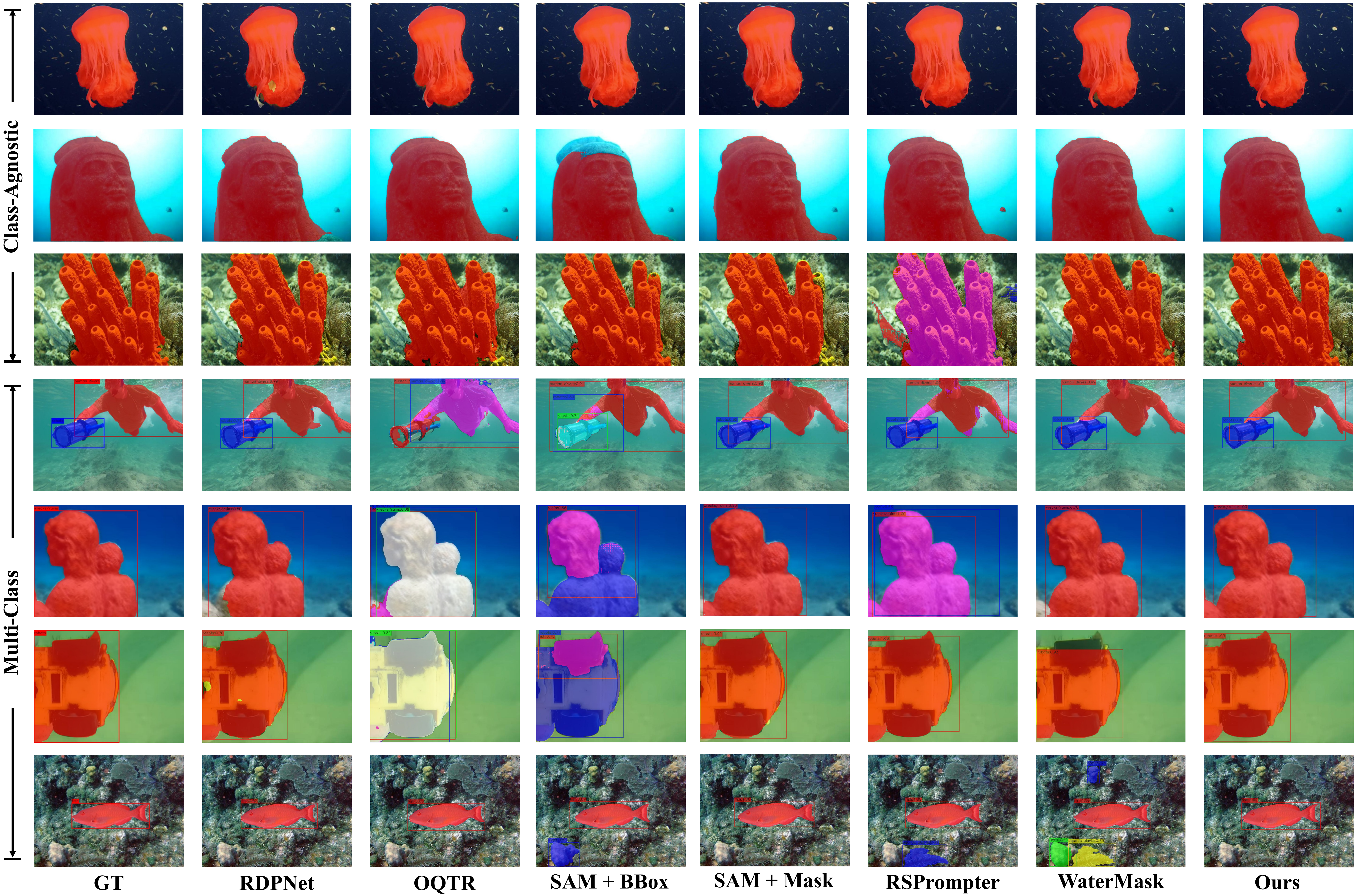}
  \vspace{-5mm}
  \caption{More qualitative comparison on \datasetName~dataset.  
  Each salient instance is represented by a unique color, and the segmented mask is superimposed on the image.    
  SAM+BBox uses inference results from Faster RCNN\cite{FasterRCNN_2015_NIPS}~as prompts for prediction, SAM+Mask stands for Mask RCNN networks \cite{MASK_RCNN_2017}~use SAM as the backbone. 
  The RSPrompter in the table is the RSPrompter-anchor framework.}
  \vspace{-2mm}
  \label{fig:supp_qc}
\end{figure*}

We also show more visual comparisons on \datasetName~dataset in \Cref{fig:supp_qc}~to demonstrate the effectiveness of our model.
It can be seen that compared to salient instance segmentation networks such as RDPNet \cite{RDPnet_2021_TIP}~and OQTR \cite{OQTR_2022_TOM}, benefiting from our Underwater Adaptive ViT Encoder, USIS-SAM can better adapt to underwater environments and segment the accurate instance boundaries (as shown in \cref{fig:supp_qc}, rows 1 and 3). And compared with instance segmentation methods such as WaterMask \cite{Lian_2023_ICCV}, RSPrompter \cite{chen2023rsprompter}, Salient Feature Prompt Generator brings USIS-SAM a more powerful salient region localization capability, which can better identify the instance's saliency (as shown in \cref{fig:supp_qc}, row 6).

\subsection{More Ablation Experiments}
\label{subsec:more_ablation}

In this subsection, we perform more detailed ablation experiments on each component in the Underwater Adaptive ViT Encoder and Salient Feature Prompt Generator.

\noindent\textbf{Number of parameters for USIS-SAM.}
In order to minimize the training consumption, we do not tune all the parameters in the original SAM architecture, but freeze the pre-trained SAM weights and insert some simple but effective modules at specific locations in the SAN to modify the SAM for underwater salience instance segmentation.
We show the number of parameters for each component in USIS-SAM in \Cref{tab:params}.
It can be seen that the trainable parameters of \methodName~constitute only 9.3\% of the total.
This means that instead of retraining or fine-tuning SAM, we can quickly apply \methodName~to underwater tasks by simply freezing SAM and specifically training our additional introduced components.

\begin{table}[!ht]
    \vskip -0.1in
    \caption{Number of parameters for each component of USIS-SAM, where $^*$~stands for trainable parameters. }
    \vskip 0.05in
    \begin{center}
    \renewcommand{\arraystretch}{1.2}
    {\begin{tabular}{c|ccc}
    \hline\hline
    Modules & Segment Anything Model& UA-ViT$^*$ & SPFG$^*$  \\
    \hline
    Params & 641.0 M & 30.5 M & 29.5 M\\
    \hline\hline
    \end{tabular}}
    \end{center}
    \vskip -0.05in\label{tab:params}
\end{table}

\noindent\textbf{Underwater Adaptive ViT Encoder.}
We show the results of the following ablation experiments about the Underwater Adaptive ViT Encoder in \Cref{tab:ab_uavit}.
First, we assess the effect of the Adapter by removing it from the UA-ViT.
The results indicate that the absence of the Adapter prevents the network from learning underwater-specific visual prompts due to the freezing of the image encoder, resulting in performance degradation of 1.4, 1.7, and 1.2 AP on the \mAP, \AP{50}, and \AP{75}, respectively.
Additionally, we also removed the Channel Adapter to demonstrate its effectiveness.
It can be seen that after removing it, the network performance decreases by 1.1, 1.3 AP, and 1.4 AP on \mAP, \AP{50}, and \AP{75}. 
This decline is attributed to the network's inability to accurately identify the importance of each channel at this time.

\begin{table}[!ht]
    \vskip -0.1in
    \caption{Effectiveness of each component in Underwater Adaptive ViT Encoder, w/o Adapter and w/o CA denote the removal of Adapters and Channel Adapter.}
    \vskip 0.05in
    \begin{center}
    \renewcommand{\arraystretch}{1.1}
    {\begin{tabular}{c|ccc}
    \hline\hline
    Methods & \mAP & \AP{50} & \AP{75} \\
    \hline
    Full Model & 43.1 & 59.0 & 48.5\\
    w/o Adapter & 41.7 (-1.4) & 57.3 (-1.7) & 47.3 (-1.2)\\
    w/o CA & 42.0 (-1.1) & 57.7 (-1.3) & 47.1 (-1.4) \\
    \hline\hline
    \end{tabular}}
    \end{center}
    \vskip -0.05in\label{tab:ab_uavit}
\end{table}

\noindent\textbf{Salient Feature Prompt Generator.}
\Cref{tab:ab_SFPG}~shows the results of ablation experiments concerning the Salient Feature Prompt Generator.
First, we evaluate the effectiveness of the Salient Feature Fusion Module by removing it. 
The results demonstrate that upon removing it, the model is unable to effectively aggregate the multilayer features of ViT, which affects its ability to accurately localize the salient regions in the image, resulting in a decrease in the AP values of \mAP, \AP{50} and \AP{75} by 0.8, 0.5 and 1.3, respectively.
In addition, we explored the effect of replacing multi-scale convolution with simple $3 \times  3$~convolution after upsampling in the Salient Feature Prompt Generator.
After replacement, the segmentation accuracy of the model decreased due to the reduced receptive field, decreasing by 0.6, 0.4, and 0.8 AP in \mAP, \AP{50}, and \AP{75}, respectively.

\begin{table}[!ht]
    \vskip -0.1in
    \caption{Effectiveness of each component in Salient Feature Prompt Generator, w/o SFFM and w/o Multi-Conv denote the removal of the salient feature fusion module and multi-up sampling module.}
    \vskip 0.05in
    \begin{center}
    \renewcommand{\arraystretch}{1.1}
    {\begin{tabular}{c|ccc}
    \hline\hline
    Methods & \mAP & \AP{50} & \AP{75} \\
    \hline
    Full Model & 43.1 & 59.0 & 48.5\\
    w/o SFFM & 42.3 (-0.8) & 58.5 (-0.5) & 47.2 (-1.3)\\
    w/o Multi-Up & 42.5 (-0.6) & 58.6 (-0.4) & 47.7 (-0.8)\\
    \hline\hline
    \end{tabular}}
    \end{center}\label{tab:ab_SFPG}
\end{table}

\noindent\textbf{Generalization Ability in similar tasks.}
We also conducted experiments on the underwater instance segmentation task, which is similar to the underwater salient instance segmentation task, to validate the generalization ability of our model.
To do so, we retrain our model on the underwater instance segmentation dataset UIIS \cite{Lian_2023_ICCV}, and compare it to other state-of-the-arts underwater instance segmentation algorithms in \Cref{tab:uiis}.
It can be seen that compared to WaterMask \cite{Lian_2023_ICCV}, which is designed for underwater instance segmentation tasks, USIS-SAM obtains a lead of 2.2 AP, 1.3 AP and 3.0 AP on mAP, AP50 and AP75, respectively.
This is attributed to the underwater vision knowledge that UA-ViT introduces to the network and the strong segmentation capability of SAM, which helps USIS-SAM to achieve favorable capabilities in similar tasks.

\begin{table}[!ht]
    \vskip -0.1in
    \caption{Quantitative comparisons with state-of-the-arts methods on underwater instance segmentation dataset UIIS \cite{Lian_2023_ICCV}. The hyperparameters and other settings for USIS-SAM in the table are the same as in \Cref{tab:usis.comp}.}
    \vskip 0.05in
    \begin{center}
    \renewcommand{\arraystretch}{1.1}
    {\begin{tabular}{c|c|ccc}
    \hline\hline
    Methods & Epoch &\mAP &\AP{50} & \AP{75} \\
    \hline
    Mask RCNN \cite{MASK_RCNN_2017}	& 36&23.4	&40.9	&25.3\\
    Point Rend	\cite{kirillov2020pointrend} &36&	25.9	&43.4&	27.6\\
    QueryInst \cite{Fang2021QuerInst}&36&	26.0	&42.8	&27.3\\
    Mask2Former \cite{Cheng2022Mask2Former} &36&	25.7	&38.0	&27.7\\
    RDPNet \cite{RDPnet_2021_TIP} &50&20.6	&38.7	&19.4\\
    WaterMask \cite{Lian_2023_ICCV}	&36&27.2	&43.7	&29.3\\
    USIS-SAM &24&	\textbf{29.4}	&\textbf{45.0}	&\textbf{32.3}\\

    \hline\hline
    \end{tabular}}
    \end{center}\label{tab:uiis}
\end{table}

\end{document}